\documentclass[runningheads]{llncs}
\usepackage[T1]{fontenc}
\usepackage{graphicx,verbatim}
\usepackage{colortbl} 
\usepackage{multirow} 
\usepackage{xcolor}   
\usepackage{array}    
\usepackage{graphicx}
\usepackage{caption}
\usepackage{algorithm}
\usepackage{algpseudocode}
\usepackage{amssymb}
\usepackage{placeins}
\usepackage{amsmath}
\usepackage{hyperref}   
\usepackage[nameinlink]{cleveref}
\usepackage{xcolor}
\usepackage{booktabs}

\newcommand{\etal}{\textit{et~al.}}
\begin{document}
\title{Pathologist Attention–Aligned Report Generation for Prostate Histopathology}
\titlerunning{Attention-Aligned Report Generation}
%

\author{Ruoyu Xue\inst{1} \and
Suryakant Singh\inst{2}$^{\dagger}$ \and 
Souradeep Chakraborty\inst{1}$^{\dagger}$ \and
Pierre Marza\inst{3,4} \and
Oksana Yaskiv\inst{5} \and
Constantin Friedman\inst{5} \and
Natallia Sheuka\inst{5} \and
Paul Friedman\inst{5} \and
Bharat Ramlal\inst{5} \and
Beatrice Knudsen\inst{6} \and
Rajarsi Gupta\inst{2} \and
Joel Saltz\inst{2} \and
Prateek Prasanna\inst{2} \and
Gregory Zelinsky\inst{7} \and
Dimitris Samaras\inst{1}}

\authorrunning{Xue et al.}
%
\institute{Department of Computer Science, Stony Brook University, USA \\
\email{\{ruoxue, souchakrabor, samaras\}@cs.stonybrook.edu}
\and
Department of Biomedical Informatics, Stony Brook University, USA \\
\email{\{suryakant.singh, Rajarsi.Gupta, Joel.Saltz, Prateek.Prasanna\}@stonybrook.edu}
\and
Université Paris-Saclay, CentraleSupélec, Gustave Roussy, INSERM, IHU PRISM, Cancer Data Science Unit, France \and Université Paris-Saclay, CentraleSupélec, MICS Laboratory, France
\email{\{pierre.marza\}@centralesupelec.fr}
\and
Department of Pathology and Laboratory Medicine, Northwell Health Laboratories \\
\email{\{OYaskiv, cfriedman7, nsheuka, pfriedman1, bramlal\}@northwell.edu}
\and
Department of Pathology, University of Utah School of Medicine \\
\email{\{beatrice.knudsen\}@path.utah.edu}
\and
Department of Psychology, Stony Brook University
\email{\{gregory.zelinsky\}@stonybrook.edu}\\
\textsuperscript{\textdagger} Equal contribution.
}


\def\mA{\mathcal{A}}
\def\mB{\mathcal{B}}
\def\mC{\mathcal{C}}
\def\mD{\mathcal{D}}
\def\mE{\mathcal{E}}
\def\mF{\mathcal{F}}
\def\mG{\mathcal{G}}
\def\mH{\mathcal{H}}
\def\mI{\mathcal{I}}
\def\mJ{\mathcal{J}}
\def\mK{\mathcal{K}}
\def\mL{\mathcal{L}}
\def\mM{\mathcal{M}}
\def\mN{\mathcal{N}}
\def\mO{\mathcal{O}}
\def\mP{\mathcal{P}}
\def\mQ{\mathcal{Q}}
\def\mR{\mathcal{R}}
\def\mS{\mathcal{S}}
\def\mT{\mathcal{T}}
\def\mU{\mathcal{U}}
\def\mV{\mathcal{V}}
\def\mW{\mathcal{W}}
\def\mX{\mathcal{X}}
\def\mY{\mathcal{Y}}
\def\mZ{\mathcal{Z}} 

\def\bbN{\mathbb{N}} 
\def\bbR{\mathbb{R}} 
\def\bbP{\mathbb{P}} 
\def\bbQ{\mathbb{Q}} 
\def\bbE{\mathbb{E}}

\def\1n{\mathbf{1}_n}
\def\0{\mathbf{0}}
\def\1{\mathbf{1}}

\def\A{{\bf A}}
\def\B{{\bf B}}
\def\C{{\bf C}}
\def\D{{\bf D}}
\def\E{{\bf E}}
\def\F{{\bf F}}
\def\G{{\bf G}}
\def\H{{\bf H}}
\def\I{{\bf I}}
\def\J{{\bf J}}
\def\K{{\bf K}}
\def\L{{\bf L}}
\def\M{{\bf M}}
\def\N{{\bf N}}
\def\O{{\bf O}}
\def\P{{\bf P}}
\def\Q{{\bf Q}}
\def\R{{\bf R}}
\def\S{{\bf S}}
\def\T{{\bf T}}
\def\U{{\bf U}}
\def\V{{\bf V}}
\def\W{{\bf W}}
\def\X{{\bf X}}
\def\Y{{\bf Y}}
\def\Z{{\bf Z}}

\def\a{{\bf a}}
\def\b{{\bf b}}
\def\c{{\bf c}}
\def\d{{\bf d}}
\def\e{{\bf e}}
\def\f{{\bf f}}
\def\g{{\bf g}}
\def\h{{\bf h}}
\def\i{{\bf i}}
\def\j{{\bf j}}
\def\k{{\bf k}}
\def\l{{\bf l}}
\def\m{{\bf m}}
\def\n{{\bf n}}
\def\o{{\bf o}}
\def\p{{\bf p}}
\def\q{{\bf q}}
\def\r{{\bf r}}
\def\s{{\bf s}}
\def\t{{\bf t}}
\def\u{{\bf u}}
\def\v{{\bf v}}
\def\w{{\bf w}}
\def\x{{\bf x}}
\def\y{{\bf y}}
\def\z{{\bf z}}

\def\balpha{\mbox{\boldmath{$\alpha$}}}
\def\bbeta{\mbox{\boldmath{$\beta$}}}
\def\bdelta{\mbox{\boldmath{$\delta$}}}
\def\bgamma{\mbox{\boldmath{$\gamma$}}}
\def\blambda{\mbox{\boldmath{$\lambda$}}}
\def\bsigma{\mbox{\boldmath{$\sigma$}}}
\def\btheta{\mbox{\boldmath{$\theta$}}}
\def\bomega{\mbox{\boldmath{$\omega$}}}
\def\bxi{\mbox{\boldmath{$\xi$}}}
\def\bnu{\mbox{\boldmath{$\nu$}}}                                  
\def\bphi{\mbox{\boldmath{$\phi$}}}
\def\bmu{\mbox{\boldmath{$\mu$}}}

\def\bDelta{\mbox{\boldmath{$\Delta$}}}
\def\bOmega{\mbox{\boldmath{$\Omega$}}}
\def\bPhi{\mbox{\boldmath{$\Phi$}}}
\def\bLambda{\mbox{\boldmath{$\Lambda$}}}
\def\bSigma{\mbox{\boldmath{$\Sigma$}}}
\def\bGamma{\mbox{\boldmath{$\Gamma$}}}
                                  
\newcommand{\myprob}[1]{\mathop{\mathbb{P}}_{#1}}

\newcommand{\myexp}[1]{\mathop{\mathbb{E}}_{#1}}

\newcommand{\mydelta}[1]{1_{#1}}

\newcommand{\myminimum}[1]{\mathop{\textrm{minimum}}_{#1}}
\newcommand{\mymaximum}[1]{\mathop{\textrm{maximum}}_{#1}}    
\newcommand{\mymin}[1]{\mathop{\textrm{minimize}}_{#1}}
\newcommand{\mymax}[1]{\mathop{\textrm{maximize}}_{#1}}
\newcommand{\mymins}[1]{\mathop{\textrm{min.}}_{#1}}
\newcommand{\mymaxs}[1]{\mathop{\textrm{max.}}_{#1}}  
\newcommand{\myargmin}[1]{\mathop{\textrm{argmin}}_{#1}} 
\newcommand{\myargmax}[1]{\mathop{\textrm{argmax}}_{#1}} 
\newcommand{\myst}{\textrm{s.t. }}

\newcommand{\denselist}{\itemsep -1pt}
\newcommand{\sparselist}{\itemsep 1pt}

\definecolor{pink}{rgb}{0.9,0.5,0.5}
\definecolor{purple}{rgb}{0.5, 0.4, 0.8}   
\definecolor{gray}{rgb}{0.3, 0.3, 0.3}
\definecolor{mygreen}{rgb}{0.2, 0.6, 0.2}

\newcommand{\cyan}[1]{\textcolor{cyan}{#1}}
\newcommand{\blue}[1]{\textcolor{blue}{#1}}
\newcommand{\magenta}[1]{\textcolor{magenta}{#1}}
\newcommand{\pink}[1]{\textcolor{pink}{#1}}
\newcommand{\green}[1]{\textcolor{green}{#1}} 
\newcommand{\gray}[1]{\textcolor{gray}{#1}}    
\newcommand{\mygreen}[1]{\textcolor{mygreen}{#1}}    
\newcommand{\purple}[1]{\textcolor{purple}{#1}}       

\definecolor{greena}{rgb}{0.4, 0.5, 0.1}
\newcommand{\greena}[1]{\textcolor{greena}{#1}}

\definecolor{bluea}{rgb}{0, 0.4, 0.6}
\newcommand{\bluea}[1]{\textcolor{bluea}{#1}}
\definecolor{reda}{rgb}{0.6, 0.2, 0.1}
\newcommand{\reda}[1]{\textcolor{reda}{#1}}

\def\changemargin#1#2{\list{}{\rightmargin#2\leftmargin#1}\item[]}
\let\endchangemargin=\endlist
                                               
\newcommand{\cm}[1]{}

\newcommand{\mhoai}[1]{{\color{blue}\textbf{[MH: #1]}}}

\newcommand{\myheading}[1]{\vspace{0.5ex}\noindent \textbf{#1}}
\newcommand{\htimesw}[2]{\mbox{$#1$$\times$$#2$}}

\newif\ifdraft
\drafttrue 
\definecolor{darkpink}{rgb}{0.91, 0.33, 0.5}
\definecolor{darkgreen}{rgb}{0.0, 0.5, 0.0}
\ifdraft
  \newcommand{\HL}[1]{{\color{orange}{\bf HL: #1}}} 
 \newcommand{\hl}[1]{{\color{orange} #1}}
 \newcommand{\RX}[1]{{\color{darkgreen}\textbf{[RX: #1]}}}
\else
  \newcommand{\HL}[1]{}
 \newcommand{\hl}[1]{#1}
 \newcommand{\ME}[1]{}
  \newcommand{\me}[1]{#1}
\fi

\newcommand{\parag}[1]{\vspace{-3mm}\paragraph{#1}}


%
%
%

\newcommand{\Sref}[1]{Sec.~\ref{#1}}
\newcommand{\Eref}[1]{Eq.~(\ref{#1})}
\newcommand{\Fref}[1]{Fig.~\ref{#1}}
\newcommand{\Tref}[1]{Table~\ref{#1}}

\newcolumntype{C}[1]{>{\centering\arraybackslash}p{#1}}

 \newcommand{\TODO}[1]{{\color{red} #1}}
\maketitle             

\begin{abstract}
The allocation of visual attention by pathologists during cancer diagnosis is a highly selective process that critically shapes the information extracted from whole-slide images (WSIs). Human attention helps medical imaging tasks such as classification and segmentation, and becomes a strong semantic cue for identifying diagnostically informative regions for report generation. In this paper, we introduce human attention into the training of pathologist report generation models. To this end, we collected a multimodal human-attention dataset of 121 prostate WSIs annotated with pathologists’ multi-scale viewport trajectories synchronized with the pathologists' verbal descriptions and cursor movements for five clinically relevant components (e.g., Gleason patterns). Using this dataset, we fine-tune two report generation models with an attention-alignment loss that regularizes the model attention over image patches to match the distribution of pathologist attention. We evaluate our approach on prostate cancer report generation and visual question answering using two models with different internal attention mechanisms (i.e., how image tokens are integrated into the language decoder). Experiments show average gains of 10.9\% on NLP-based metrics and 19.3\% in accuracy across five clinically relevant report components. Further, model attention maps extracted at inference time, with minimal computational overhead, align more closely with pathologist attention, providing stronger visual support for the generated reports by highlighting the regions that most influence the output.



\end{abstract}
\section{Introduction}
\begin{figure}[t!]
  \centering
  \includegraphics[trim={0 0 0.3cm 0},width=1.0\linewidth]{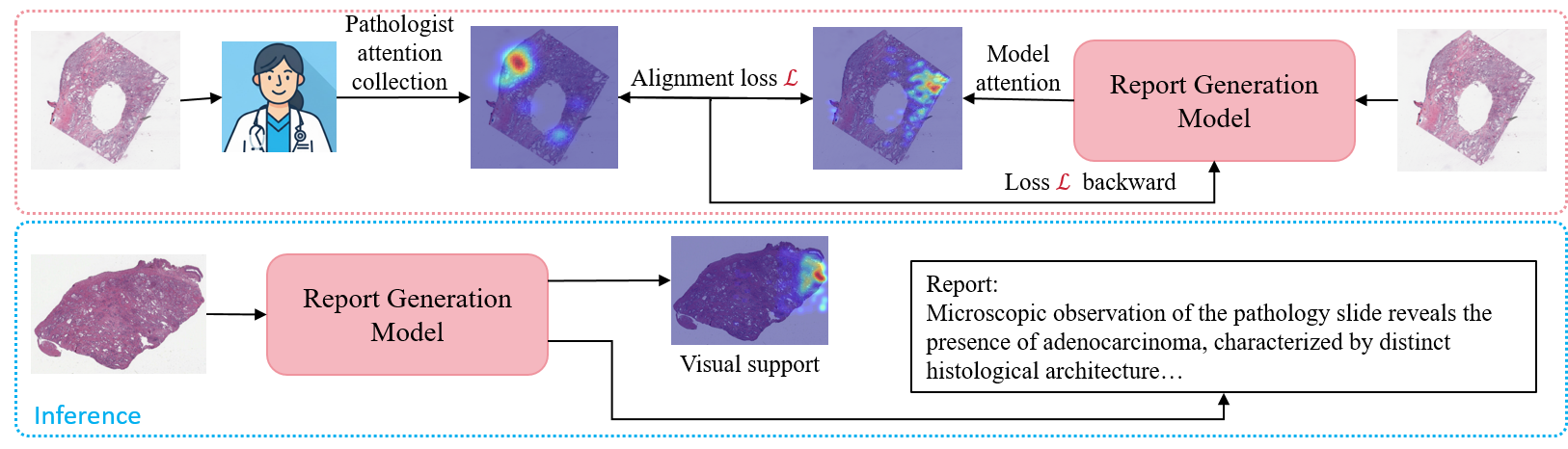}
  \caption{Top-left: Collect pathologist attention during WSI examination. Top-right: Align model attention maps to human attention via a training-time loss. Bottom: At inference, the model generates reports with model-attention-based visual evidence.}
  \label{fig:teaser}
\end{figure}


Human visual attention is capacity-limited and shifts over time. During whole-slide image (WSI) examination, pathologists scan the slide sequentially, selectively focusing on task-relevant regions. These attention trajectories reflect the visual evidence a pathologist encodes during cancer diagnosis. Clinically, modeling pathologist attention can support trainee education by revealing how specialists examine slides and can also facilitate the assessment of pathologists’ expertise levels~\cite{chakraborty2025measuring,chakraborty2022predicting,brunye2014eye,chakraborty2024decoding}, and understand decision-making \cite{thai2025eye}. Computationally, expert attention has increasingly been used as auxiliary supervision in medical imaging. Across pathology and radiology, attention signals have guided fully supervised and weakly supervised classification \cite{nan2025deep,bhattacharya2025gazelt,bhattacharya2024gazediff}, segmentation \cite{ge2025adaptation}, and text-to-image generation \cite{bhattacharya2024radgazegen}. Beyond medical imaging, evidence from cognitive science and attention-aware image description \cite{griffin2000eyes,coco2012scan,xue2025personalized} suggests that where people look shapes how they interpret and describe visual content. Together, these findings motivate leveraging pathologists’ attention as an auxiliary cue for pathology report generation,
which grounds generated text in plausible visual evidence, and improves both interpretability and report quality.\\
Despite rapid recent progress in pathology report generation and visual question answering~\cite{liang2025wsi,guo2024histgen,chen2024wsicaption,zhang2025historical,tran2025generating}, only a few works incorporate auxiliary supervision such as model-predicted semantic information, and none so far leverage human attention in pathology report generation. Kim \etal~\cite{kim2025enhancing} introduces semantic cues via a multi-label classifier over key diagnostic elements, but the generated reports are limited to predefined categories and depend on predicted semantics rather than ground-truth signals. Agent-based approaches such as CPathAgent~\cite{sun2025cpathagent} and PathChat+~\cite{chen2025evidence} supervise the training by predicting the Region Of Interest(ROI)-based visual support with step-by-step reasoning; however, these ROIs are not derived from the attention of actual pathologists, and inducing such visual support typically requires large-scale instruction tuning.\\
In this paper, we align the visual attention of report generation models with pathologist attention. In vision–language transformers, model attention weights indicate how much each image patch contributes to token generation, providing a natural mechanism for grounding diagnostic statements in visual evidence. To supervise this alignment, we first collect a prostate cancer pathologist-attention dataset that captures multi-scale WSI examination behavior for five clinically critical components: Gleason patterns, perineural invasion, intraductal carcinoma, extraprostatic extension, and surgical margin. While the pathologists read the slides, we record viewport trajectories across magnification levels and synchronized, real-time verbal descriptions of observed findings, which provide soft labels for all the five components across the corresponding viewports.\\
Using this rich supervision, we introduce a plug-and-play module for report generation in the form of an auxiliary attention-alignment loss that regularizes model attention toward the corresponding pathologist attention distribution over image tokens (\Cref{fig:teaser}). Pathologist attention is required only during training, with no additional inputs at inference, and the loss adds negligible overhead. We evaluate this design on both a report generation model and a WSI visual question answering model, covering two common vision–language fusion designs: cross-attention and self-attention. Experiments show 5.5\%–12\% average improvements on NLP-based metrics, and 8\%–34\% gains in accuracy on the five clinically related components across the two models. The model attention in the inference-time is aligned more closely with pathologist attention, providing stronger visual evidence for the generated reports. Our contributions are:
\begin{itemize}
\item To our knowledge, we are the first to incorporate pathologist-attention supervision into pathology report generation, showing consistent report-quality gains across architectures.
\item We collect a multimodal pathologist-attention dataset of prostate WSIs with multi-scale viewport trajectories synchronized with spoken descriptions of observed findings.
\item We improve interpretability by producing attention-based visual evidence that better matches how pathologists attend during cancer diagnosis.
\end{itemize}
As we show, fine-tuning a pathology report generation model with human attention improves performance with relatively small pathologist annotation effort ($\sim7$ hours in total for the reported experiments). This suggests that our plug-and-play module can easily become a part of model training for pathology report generation. 
\section{Pathologist Attention Dataset} 
\begin{figure}[t!]
  \centering
  \includegraphics[trim={0 0 0.3cm 0},width=1.0\linewidth]{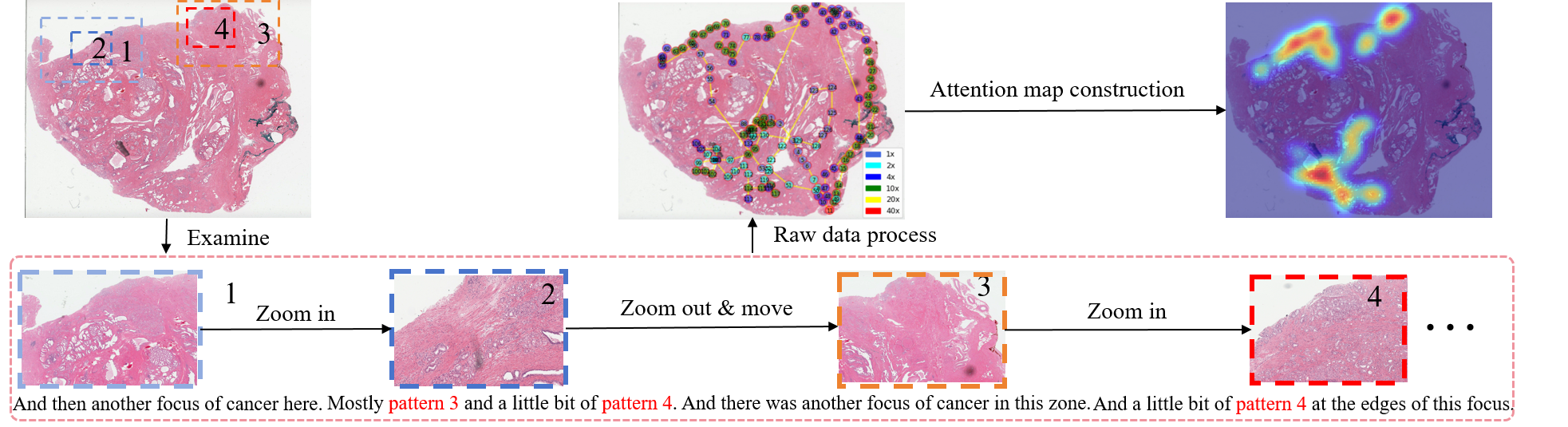}
  \caption{\textbf{Data collection process}. Pathologists examine each WSI by iteratively zooming for detail and zooming out to navigate (bottom right). We record synchronized verbal descriptions, convert viewports center to scanpath (top center), filter points by target components, and generate a Gaussian-smoothed saliency map overlaid on the slide (top right).}
  \label{fig:data-collection}
\end{figure}

\textbf{Dataset collection.} 
We record pathologists’ attention using QuIP caMicroscope~\cite{saltz2017containerized}. Six pathologists 
reviewed 121 TCGA-PRAD prostate WSIs~\cite{zuley2016radiology} and annotated seven elements: Gleason score, intraductal carcinoma (IDC), perineural invasion (PNI), extraprostatic extension (EPE), surgical margin, lymphovascular invasion, and seminal vesicle invasion. As shown in \Cref{fig:data-collection}, slides are viewed in a 1912$\times$930 viewport with iterative zooming and panning, yielding multi-magnification viewports. We sample viewports at 20,Hz and convert each to its center point to form a magnification-aware scanpath; synchronized verbal descriptions provide soft component labels. We remove duplicates and keep only positively labeled points per component. Due to sparsity, we exclude lymphovascular and seminal vesicle invasion and focus on five components. On average, scanpaths contain 59 points with 221,s per slide ($\sim7$ hours total).\\
\textbf{Human attention map construction.} We construct Gaussian-blurred saliency maps~\cite{jiang2015salicon,chakraborty2022predicting,chakraborty2025measuring}. 
For each magnification ($2\times$, $4\times$, $10\times$, $20\times$), we project scanpath points onto the corresponding grid, apply Gaussian smoothing, then resize and merge maps into a single multi-scale saliency map per slide.
\section{Method}
\subsection{Preliminaries}
Report generation models adopt transformers structure~\cite{vaswani2017attention}. 
Given queries $Q$, keys $K$, and values $V$, scaled dot-product attention computes
\begin{equation}
A=\mathrm{softmax}\!\left(\frac{QK^\top}{\sqrt{d}}\right), \qquad 
\mathrm{Attn}(Q,K,V)=AV,
\end{equation}
where $A$ is the model attention weights, and $\mathrm{Attn}(Q,K,V)$ is the output feature of a transformer attention layer. \\
Multimodal report generation models typically use two image-text fusion designs. 
\textbf{Self-attention fusion} concatenates image and text tokens $Z=[\mathbf{X};\mathbf{Y}_{<t}]$, where $\mathbf{X}\in\mathbb{R}^{I\times d}$ are image tokens and $\mathbf{Y}_{<t}\in\mathbb{R}^{t\times d}$ are previously generated text tokens, and projects $Q,K,V$ from $Z$. 
\textbf{Cross-attention fusion} projects $Q$ from the current text query, while $K,V$ come only from $\mathbf{X}$. 
These fusion determine how slide features influence next-token prediction: WSI-LLaVA~\cite{liang2025wsi} uses self-attention fusion, whereas HistGen~\cite{guo2024histgen}, Bi-Gen~\cite{zhang2025historical}, and HistoGPT~\cite{tran2025generating} adopt cross-attention.

\subsection{Pathologist Attention Alignment}
\label{sec:human_encode}
\textbf{Pathologist attention encoding in token space. }Pathology report generation models first encode each WSI into a sequence of image tokens $\mathbf{F}\in\mathbb{R}^{I\times d}$. Given a full-resolution human saliency map, we assign each token a saliency value by querying the map at the token’s recorded patch coordinates, yielding $\mathbf{a}^{\text{human}}\in\mathbb{R}^{I}$. 
To encourage model attention consistent with pathologists' attention, we fine-tune WSI-LLaVA and HistGen with the standard autoregressive next-token objective augmented by an attention-alignment loss. \\
\textbf{Human attention alignment in WSI-LLaVA \cite{liang2025wsi}}. We use the last-layer self-attention weights  $\mathbf{A}\in\mathbb{R}^{(I+P+T)\times(I+P+T)}$, where $I$ is the number of image tokens, $P$ is the prompt length, and $T$ is the number of answer tokens. 
We extract the answer-to-image attention block
\begin{equation}
\mathbf{A}_{\text{ans}\rightarrow \text{img}}
=\mathbf{A}\big[\mathcal{T}_{\text{ans}},\,\mathcal{I}\big]\in\mathbb{R}^{T\times I},
\end{equation}
where $\mathcal{I}$ indexes image-token positions and $\mathcal{T}_{\text{ans}}$ indexes answer-token query positions. 
We then aggregate token-wise attention into an image-level model attention vector by averaging over answer tokens:
\begin{equation}
\mathbf{a}^{\text{model}}
=\frac{1}{T}\sum_{t\in\mathcal{T}_{\text{ans}}}\mathbf{A}_{\text{ans}\rightarrow \text{img}}[t,:]\in\mathbb{R}^{I}.
\end{equation}
Given the corresponding human attention vector $\mathbf{a}^{\text{human}}\in\mathbb{R}^{I}$, we define the alignment loss as the KL divergence between the attention distributions following the design of GazeVLM \cite{pani2025gaze},
\begin{equation}
\mathcal{L}_{\mathrm{KL}}
=
D_{\mathrm{KL}}\!\left(
\mathrm{softmax}(\mathbf{a}^{\text{human}})
\ \Big\|\ 
\mathrm{softmax}(\mathbf{a}^{\text{model}})
\right).
\end{equation}

\textbf{Training objective.}
Given the ground-truth tokens $\{t_1,\dots,t_T\}$, the base objective is the next-token negative log-likelihood 
$\mathcal{L}_{\mathrm{CE}} = -\sum_{t=1}^{T} \log p_{\theta}(t_t \mid t_{<t}, \mathbf{X})$. 
The overall loss is 
$\mathcal{L} = \mathcal{L}_{\mathrm{CE}} + \lambda\,\mathcal{L}_{\mathrm{KL}}$, 
where $\lambda$ controls the strength of alignment, and is set to 1.0 in the experiment.\\
\textbf{Human attention alignment in HistGen}. Given the different attention mechanism of HistGen (cross-attention) from WSI-LLaVA (self-attention), we directly extract the model attention $\mathbf{A}^{\text{model}}\in\mathbb{R}^{T\times I}$ and then follow the same way to compute alignment loss.

\subsection{Inference-Time Visual Support Generation}

During inference, we generate reports and extract model attention to image tokens in each step, then average across steps to obtain a slide-level vector $\mathbf{a}^{\text{model}}$, then back-project it to a pixel-level heatmap by assigning each patch its token value, followed by normalization and Gaussian smoothing (~\Cref{sec:human_encode}).

\section{Experiments}
\subsection{Setups}
\textbf{Baselines. } We use HistGen \cite{guo2024histgen} and WSI-LLaVA \cite{liang2025wsi} as base models, which differ in multimodal fusion attention mechanisms and are designed for different tasks (report generation vs.\ visual question answering). \textbf{Zero-shot} denotes directly evaluating each pretrained model on its test set. \textbf{w/o alignment} denotes LoRA fine-tuning \cite{hu2022lora} using only the next-token prediction loss, without any pathologist-attention alignment.\\
\textbf{Report Dataset. }We use the official datasets and splits of HistGen and WSI-LLaVA. For HistGen, attention alignment is trained only on the 81 slides with our attention annotations, while evaluation follows the official val/test sets (86/76). Visual-support evaluation (requiring GT comparison) is performed on 20 test slides. For WSI-LLaVA, we use 107 training examples and 11 test examples for report-quality evaluation, and 15 test samples for visual-support evaluation.\\
\textbf{Metrics. } We evaluate both report quality and visual support. For report generation, we report standard NLP metrics: BLEU (B1--B4)~\cite{papineni2002bleu}, METEOR (M)~\cite{banerjee2005meteor}, and ROUGE-L (R)~\cite{lin2004rouge}. Since our attention annotations target five pathological components, we use ChatGPT-5.2 to extract them and compute exact-match scores between generated and ground-truth reports: Gleason requires matching both primary and secondary patterns, while the remaining components are binary scored. We report ACC-G, ACC-PNI, ACC-EPE, and ACC-S for Gleason score, perineural invasion, extraprostatic extension, and surgical margin, respectively; we omit intraductal carcinoma because it does not appear in the ground-truth reports. For visual support, we measure attention heatmap similarity with Normalized Scanpath Saliency (NSS) and KL divergence (KL)~\cite{peters2005components}, and quantify agreement with ground-truth Gleason regions using segmentation labels from~\cite{refined_tcga_prad_dataset_hf}, reporting ROC-AUC (the probability that a pixel inside the binary segmentation mask receives a higher heatmap score than a pixel outside).\\
\textbf{Implementation details.} We use LoRA finetuning for both HistGen and WSI-LLaVA with the same split and training setting as the original papers. The experiment runs on a single RTX 6000 GPU. We set $\lambda=1.0$ for both models. \\

\begin{table*}[t]
\centering
\scriptsize
\resizebox{1.0\textwidth}{!}{
\begin{tabular}{c|c|cccccc|cccc}
\toprule
Backbone & Method & \textbf{B1} & \textbf{B2} & \textbf{B3} & \textbf{B4} & \textbf{R} & \textbf{M} &
\textbf{ACC-G} & \textbf{ACC-PNI} & \textbf{ACC-EDE} & \textbf{ACC-S}\\
\midrule
\multirow{3}{*}{HistGen~\cite{guo2024histgen}}
& Zero-shot      & 0.329 & 0.162 & 0.087 & 0.048 & 0.172 & 0.150 & 25.0 & 25.7 & 30.0 & 25.7 \\
& w/o Alignment  & 0.326 & 0.174 & 0.102 & 0.065 & 0.185 & 0.151 & 30.0 & 35.0  & 34.3 & 28.5\\
& Ours           & \textbf{0.361} & \textbf{0.194} & \textbf{0.116} & \textbf{0.077} & \textbf{0.203} & \textbf{0.166} & \textbf{35.7} & \textbf{40.0}  & \textbf{44.3} & \textbf{30.0} \\
\midrule
\multirow{3}{*}{\shortstack{WSI-LLaVA\\\cite{liang2025wsi}}}
& Zero-shot      & 0.469 & 0.283 & 0.187 & 0.129 & 0.328 & 0.211 & 45.5 & 81.8 & 45.5 & -\\
& w/o Alignment  & 0.444 & 0.254 & 0.146 & 0.093 & 0.300 & 0.220 & 54.5 & 81.8 & 45.5 & -\\
& Ours           & \textbf{0.479} & \textbf{0.292} & \textbf{0.197} & \textbf{0.136} & \textbf{0.338} & \textbf{0.228} & \textbf{63.6} & \textbf{90.9} & \textbf{63.6} & -\\
\bottomrule
\end{tabular}
}
\caption{Quantitative results on the test set of pathologist-aligned report generation performance on prostate cancer. ACC-S of WSI-LLaVA is not reported due to missing annotations in the ground-truth reports.\label{tab:main-result}}
\end{table*}
\begin{figure}[t!]
  \centering
  \includegraphics[trim={0 0 0.3cm 0},width=1.0\linewidth]{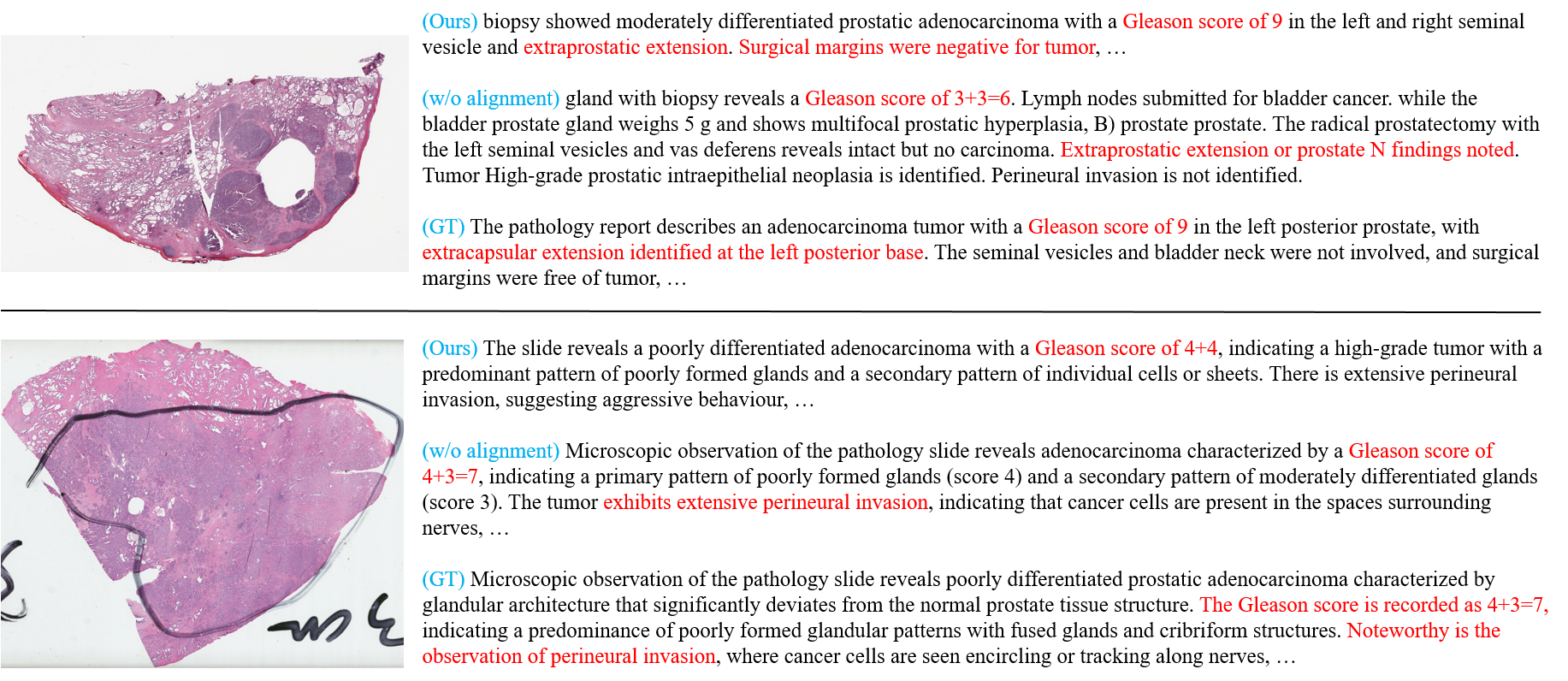}
  \caption{\textbf{Qualitative results of report generation. }Top: HistGen; Bottom: WSI-LLaVA. Components mentioned in the generated reports are highlighted in \textcolor{red}{red}, indicating more complete and higher-quality reports. }
  \label{fig:qual-main-result}
\end{figure}

\begin{figure}[t!]
  \centering
  \includegraphics[trim={0 0 0.3cm 0},width=1.0\linewidth]{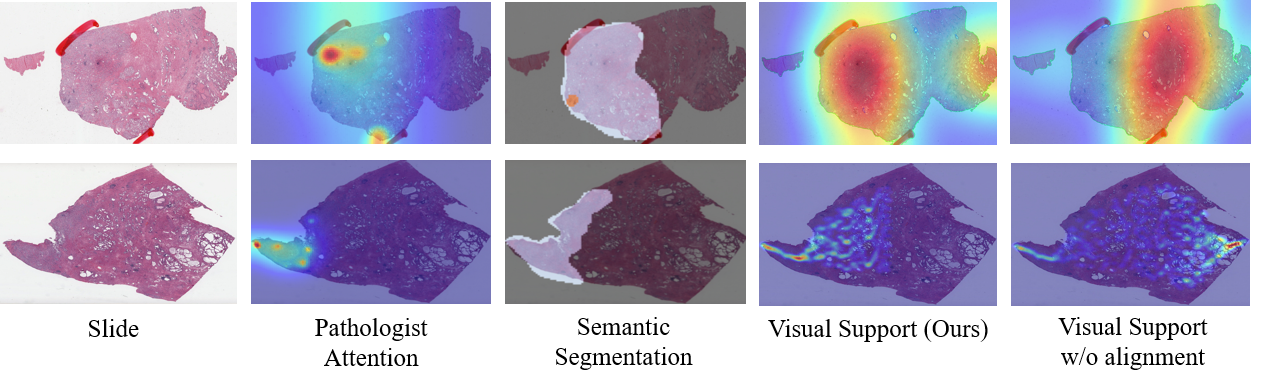}
    \caption{\textbf{Qualitative results of visual support.} Visual support is extracted from the last decoder layer at inference (top: HistGen, blurred due to the small number of image tokens for better visualization; bottom: WSI-LLaVA). Compared to training without alignment (fifth column), our method (fourth column) suppresses non-Gleason regions and better matches ground-truth pathologist attention (second column) and Gleason segmentation \cite{refined_tcga_prad_dataset_hf} (third column).}
  \label{fig:quality}
\end{figure}

\begin{table}[t]
\centering
\footnotesize
\setlength{\tabcolsep}{3.5pt}
\renewcommand{\arraystretch}{1.05}
\begin{tabular}{c|ccc|ccc}
\hline
\multirow{2}{*}{\textbf{Method}} 
& \multicolumn{3}{c|}{\textbf{HistGen}} 
& \multicolumn{3}{c}{\textbf{WSI-LLaVA}} \\
\cline{2-4}\cline{5-7}
& NSS$\uparrow$ & KL$\downarrow$ & ROC-AUC $\uparrow$
& NSS$\uparrow$ & KL$\downarrow$ & ROC-AUC $\uparrow$ \\
\hline
w/o alignment & 0.43 & 5.45 & 0.500 & 0.76 & 3.10 & 0.598 \\
with alignment (Ours) & \textbf{0.59} & \textbf{4.76} & \textbf{0.537} & \textbf{0.87} & \textbf{2.98} & \textbf{0.615} \\
\hline
\end{tabular}
\caption{\textbf{Quantitative results of visual support. }Our method produces model attention that aligns better with human and segmentation labels.}
\label{tab:visual-evidence}
\end{table}
\begin{table}[t]
\centering
\footnotesize
\setlength{\tabcolsep}{3.5pt}
\renewcommand{\arraystretch}{1.05}
\begin{tabular}{c|ccc|ccc}
\hline
\multirow{2}{*}{\textbf{Method}} 
& \multicolumn{3}{c|}{\textbf{HistGen}} 
& \multicolumn{3}{c}{\textbf{WSI-LLaVA}} \\
\cline{2-4}\cline{5-7}
& B4 & R & ACC-G 
& B4 & R & ACC-G \\
\hline
all layers & \textbf{0.079} & 0.199 & 34.2 & 0.124 & 0.325 & 45.5 \\
last layer (Ours) & 0.077 & \textbf{0.203} & \textbf{35.7} & \textbf{0.136} & \textbf{0.338} & \textbf{63.6} \\
\hline
\end{tabular}
\caption{\textbf{Ablation on attention-extraction layer.} Applying the attention-alignment loss at every layer yields similar or worse performance than supervising only the last layer, supporting our choice of a lightweight last-layer design with minimal overhead.}
\label{tab:layer-ablation}
\vskip -0.2in  
\end{table}
\subsection{Results}
\textbf{Quantitative results.} In \Cref{tab:main-result}, aligning pathologist attention achieves the best performance on both HistGen and WSI-LLaVA. In particular, our method yields higher accuracy on all five clinical components than the baselines, suggesting that human attention guides the model toward more informative regions. We also observe that fine-tuning WSI-LLaVA without attention alignment leads to a performance drop, likely due to overfitting in this low-data setting, as WSI-LLaVA is built on a large language model (Vicuna-7b-v1.5~\cite{zheng2023judging}). By providing human attention as a visual constraint, this alignment discourages the model from memorizing report text.\\
\textbf{Qualitative results of generated reports. } In \Cref{fig:qual-main-result}, our model captures the clinically critical components targeted in our dataset collection more consistently than the baselines, suggesting it learns to leverage pathologist attention cues effectively. \\
\textbf{Visual support.} We report qualitative and quantitative results on visual support in \Cref{fig:quality} and \Cref{tab:visual-evidence}, evaluating (1) alignment with pathologist attention and (2) coverage of Gleason patterns. These results suggest our alignment loss encourages the model to focus on more informative, human-preferred regions, improving report quality. The resulting visual support also enhances interpretability by indicating which regions support the model’s diagnostic statements.\\
\textbf{Ablation.} We ablate which transformer layer to use for extracting model attention in human-attention alignment. Results in \Cref{tab:layer-ablation} show that supervising only the last layer is sufficient to steer the model toward pathologist-attended regions, likely because last-layer attention most directly influences next-token prediction. In contrast, aligning multiple layers in WSI-LLaVA degrades performance below the zero-shot baseline, likely because it is much deeper than HistGen due to its large language model backbone, where different layers encode distinct information and enforcing a uniform attention target across all layers is overly restrictive.
\begin{table}[t]
\centering
\footnotesize
\setlength{\tabcolsep}{3.5pt}
\renewcommand{\arraystretch}{1.05}
\begin{tabular}{c|ccc|ccc}
\hline
\multirow{2}{*}{\textbf{Method}} 
& \multicolumn{3}{c|}{\textbf{HistGen}} 
& \multicolumn{3}{c}{\textbf{WSI-LLaVA}} \\
\cline{2-4}\cline{5-7}
& B4 & R & ACC-G 
& B4 & R & ACC-G \\
\hline
all layers & \textbf{0.079} & 0.199 & 34.2 & 0.124 & 0.325 & 45.5 \\
last layer (Ours) & 0.077 & \textbf{0.203} & \textbf{35.7} & \textbf{0.136} & \textbf{0.338} & \textbf{63.6} \\
\hline
\end{tabular}
\caption{\textbf{Ablation on attention-extraction layer.} Applying the attention-alignment loss at every layer yields similar or worse performance than supervising only the last layer, supporting our choice of a lightweight last-layer design with minimal overhead.}
\label{tab:layer-ablation}
\vskip -0.2in  
\end{table}
\section{Conclusion and Discussion}
We propose a plug-and-play pathologist attention–alignment framework that uses multi-scale expert attention as auxiliary supervision for explainable pathology report generation. Across HistGen and WSI-LLaVA, it improves report quality and clinically relevant component accuracy while producing attention maps that better match pathologists’ attention with minimal overhead. Our results highlight the value of auxiliary signals for report generation and suggest scalability to additional cancer types due to the modest data-collection burden. Future work will expand attention collection to more cancer types and develop models that generalize to new cancer types by leveraging shared examination behaviors of pathologists among different cancer types.

\section{Acknowledgement}
This research is supported by US National Science Foundation (NSF) grants IIS-2123920, IIS-2212046 and 2442053, National Institutes
of Health (NIH), National Cancer Institute (NCI) grants
1R21CA25849301A1, 1R01CA297843-01, 3R21CA25849302S1, 1R03DE033489-01A1 and the \textit{Health Data
Hub} as part of the second edition of the \textit{France-Québec} call for projects \textit{Intelligence Artificielle en santé}. The content is solely the responsibility of the authors and does not necessarily represent the official views of the National Institutes of Health.

\noindent\textbf{Disclosure of Interests.} The authors have no competing interests to declare that are relevant to the content of this article. 

%
%
%
\bibliographystyle{splncs04}
\bibliography{mybib}
%




\end{document}